\documentclass[runningheads]{llncs}

\usepackage{eccv}

\usepackage{eccvabbrv}

\usepackage{graphicx}
\usepackage{booktabs}

\usepackage[accsupp]{axessibility}  

\usepackage{graphicx}

\usepackage{tikz}
\usepackage{comment}
\usepackage{amsmath,amssymb} 
\usepackage{bbm}
\usepackage{booktabs}
\usepackage[colorlinks]{hyperref}
\usepackage{graphicx}
\usepackage{subfig}
\usepackage{bm}
\usepackage{multirow}
\usepackage{wrapfig}

\usepackage{hyperref}

\usepackage{orcidlink}

\usepackage{amstext}
\usepackage{amssymb}
\usepackage{booktabs}

\usepackage{tikz}
\usepackage{comment}
\usepackage{amsmath,amssymb} 
\usepackage{bbm}
\usepackage{booktabs}
\usepackage{graphicx}
\usepackage{bm}
\usepackage{multirow}
\usepackage{cases}
\usepackage{color}
\usepackage{ulem}

\usepackage{url}
\usepackage{epsfig}
\usepackage{amsmath}
\usepackage{amssymb}
\usepackage{multirow}
\usepackage{soul}
\usepackage{footnote}
\usepackage{tablefootnote}
\usepackage{caption}
\usepackage{subcaption}

\usepackage{url}
\usepackage{multirow}
\usepackage{booktabs}
\usepackage{pifont}
\usepackage{xcolor}
\usepackage{graphicx}
\usepackage{amsmath}
\usepackage{algorithm}
\usepackage{mathtools}
\usepackage{algorithm}
\usepackage{algpseudocode}
\usepackage{listings}
\usepackage{wrapfig}
\usepackage{subcaption}
\usepackage{xcolor,colortbl}

\newcommand{\Renyi}{{R\'enyi}}

\newcommand{\cmark}{\color{green}\ding{51}}%
\newcommand{\xmark}{\color{red}\ding{55}}%

\definecolor{top1}{RGB}{255,179,179}
\definecolor{top2}{RGB}{255,217,179}
\definecolor{top3}{RGB}{255,255,179}

\usepackage{tikz}
\usepackage{comment}
\usepackage{bbm}
\usepackage{booktabs}
\usepackage{graphicx}
\usepackage{bm}
\usepackage{multirow}

\makeatletter
\providecommand{\sf@counterlist}{}
\makeatother

\begin{document}

\title{
{GTP}-4o: 
Modality-prompted Heterogeneous Graph Learning for Omni-modal Biomedical Representation
} 

\titlerunning{GTP-4o: Modality-prompted Heterogeneous
Graph for Omni-modality}

\author{
Chenxin Li\inst{1} \and
Xinyu Liu\inst{1}\thanks{Equal contribution} \and
Cheng Wang\inst{1}\textsuperscript{$\star$}\and
Yifan Liu\inst{1}  \and
Weihao Yu\inst{1}  \and
Jing Shao\inst{2} \and
Yixuan Yuan\inst{1}
}

\authorrunning{C.Li et al.}

\institute{The Chinese University of Hong Kong, Hong Kong SAR. 
\and
Shanghai AI Laboratory, Shanghai, China
} 
\maketitle

\begin{abstract}
Recent advances in learning multi-modal representation have witnessed the success in biomedical domains. While established techniques enable handling multi-modal information,  the challenges are posed when extended to various clinical modalities and practical modality-missing setting due to the inherent modality gaps. To tackle these, we propose an innovative {M}odality-\underline{p}rompted H{e}\underline{t}e{r}ogeneous \underline{G}raph \underline{for} \underline{O}mni-modal L{e}arning (GTP-4o), which embeds the numerous disparate clinical modalities into a unified  representation, completes the deficient embedding of missing modality and reformulates the cross-modal learning with a graph-based aggregation. Specially, we establish a heterogeneous graph embedding to explicitly capture the diverse semantic properties on both the modality-specific features (nodes) and the cross-modal relations (edges).
Then, we design a modality-prompted completion that enables completing the inadequate graph representation of missing modality through a graph prompting mechanism, which generates hallucination graphic topologies to steer the missing embedding towards the intact representation.
Through the completed graph, we meticulously develop a knowledge-guided hierarchical cross-modal aggregation consisting of a global meta-path neighbouring to uncover the potential heterogeneous neighbors along the pathways driven by domain knowledge, and a local multi-relation aggregation module for the comprehensive cross-modal interaction across various heterogeneous relations.
We assess the efficacy of our methodology on rigorous benchmarking experiments against prior state-of-the-arts. In a nutshell, GTP-4o presents an initial foray into the intriguing realm of embedding, relating and perceiving the heterogeneous patterns from various clinical modalities holistically via a graph theory.
Project page: \url{https://gtp-4-o.github.io/}

\end{abstract}

\section{Introduction}
\label{sec:intro}


Each modality has its own perspective to reflect the specific data characteristics~\cite{zhou2019review,li2024u,li2021unsupervised,li2022domain}. Integrating multi-modal data empowers the models with various insights into the conditions of subjects at the macroscopic, microscopic, and molecular levels, and allows for an accurate and comprehensive disease diagnosis~\cite{li2020multimodaldisease,wuyang2021joint,li2023adjustment,liu2022intervention,xu2024immunotherapy}.
For instance, multimodal fusion of various imaging techniques has significantly improved gastrointestinal lesion detection and characterization in endoscopic scenes~\cite{liu2024endogaussian,li2024endora,liu2024lgs,li2024endosparse}.
Similarly, incorporating genomic information with pathological images can improve the prediction accuracy of cancer grading~\cite{richard2022tmi,chen2021multimodal,Xu_2023_ICCV,yang2023mrm,he2023h}.
A relevant task, survival prediction, which aims to predict the time interval to a significant event such as death or disease relapse, can also benefit from such multi-modal inclusion~\cite{chen2021multimodalsurvival}.
Besides, the cell graphs constructed by the cell nuclei segmentation of pathological images, are shown to provide more fine-grained microscopic information~\cite{xu2022instancecellgraph0}.
Recent advances in visual language models also sparks the works in learning from biomedical images and texts~\cite{zhang2023biomedclip}, whereby the diagnostic texts usually encapsulates abstract semantic information~\cite{ding2022unsupervised}. 
These progress presents potential for extending the capacity boundary of biomedical multi-modal models to omni-modal representation to handle a broader range of clinical modalities.

\begin{figure}[!t]   
	\centering	   
\includegraphics[width=\linewidth]{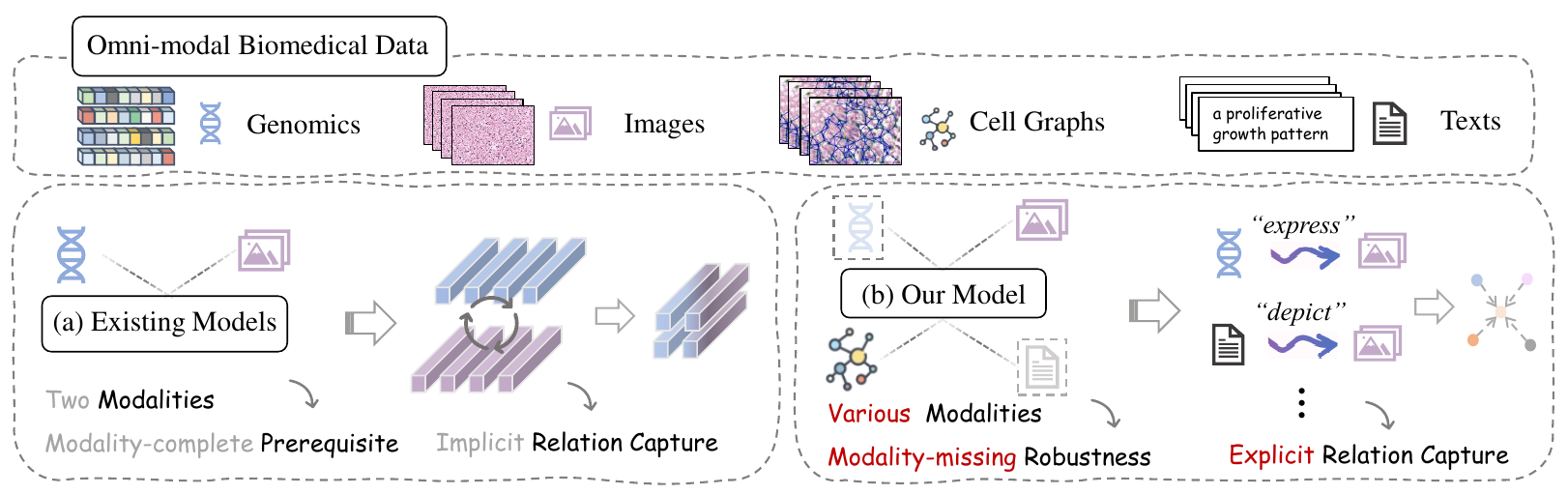}  
\caption{
\textbf{Methodology Comparison}.
Unlike \textbf{(a)} prior methods, \textbf{(b)} our framework enables learning unified omni-modal representation from various clinical modalities with modality missing and explicit capture of the cross-modal relations through the established heterogeneous graph representation.
}
  \label{fig:teaser}   
\vspace{-12Pt}
\end{figure} 


Established multimodal methods typically follow the principle that first extracts uni-modal features, then learns cross-modal relations in paired multimodal data~\cite{zhang2023biomedclip, radford2021clip, xue2023ulip,li2022sigma_da_sigma}.
Early researches design meticulous fusion techniques for multimodal information and wish to maximize the benefits of each modality~\cite{huang2020fusion,li2022scan}.
Due to the imbalanced learning process 
with inherent modal disparity and heterogeneity~\cite{zhang2020multirelation, zhang2021modalityawaremutual, kim2023heterogeneous}, recent efforts pivot towards improving collaborative learning of multiple modalities by balancing and adjusting learning of each modality~\cite{gao2019dynamicattention, xue2023dynamicfeature, peng2022balancedgradientdynamic, li2022deepfusion11, zhang2022transformer22}.
Through deriving the modality-relevant weighting factors, these methods dynamically modulates the learning and fusing of multimodal information on features~\cite{xue2023dynamicfeature}, gradients~\cite{peng2022balancedgradientdynamic}, attentions~\cite{gao2019dynamicattention,liu2024stereo}, etc.

Despite the success in alleviating the modality gap, the challenge remains severe when applied to (especially a broad range of) biomedical modalities,  primarily due to the two featured challenges.
The first challenge lies in the large semantic heterogeneity exhibiting on biological modalities. 
A straightforward example is that an \textit{``dog''} in natural images shares similar object-related semantics its sound, while the semantic relation and local correspondence between genomic profiles and pathological images is highly ambiguous~\cite{richard2022tmi,chen2021multimodal,Xu_2023_ICCV,li2023sigma++,liu2023decoupled}.
Through prior methods employ optimal transport~\cite{Xu_2023_ICCV} or cross-modal attention~\cite{chen2021multimodalsurvival,li2022scan++} to capture fine-grained correlation across genes and images, they still overlook the heterogeneity in a high-order space, i.e., relations across modalities.
Every two modalities have their own relation with specific semantics and attributes.
As shown in Fig.~\ref{fig:teaser}, the relation across images and genomics is semantically related to \textit{``express''}, 
while that across images and texts could be abstracted as \textit{``depict''}.
Therefore, these observations inspire us to introduce a unified non-Euclidean representation that explicitly captures the heterogeneous attributes on both modal features and cross-modal relations.

Secondly, in clinical practice, it is common to encounter partial absence in some modalities due to privacy and ethical considerations.
The limitations in data collection technology and the concerns surrounding bioinformatics security make it more challenging to access all the data modalities.
However, most multimodal methods have a common assumption on the data completeness~\cite{radford2021clip, zhang2023biomedclip, xue2023ulip}.
Once a modality is missing regardless of training or testing, the multimodal fusion becomes unreachable, which leads to sub-optimal performance~\cite{lee2023multimodalPromptingwithMiss}.
Therefore, we are committed to designing algorithms to adaptively complete the feature space messed up by the missing of the modality such that all the representation from the missing and existing modality could be 
handled in a unified fashion.

To address the aforementioned challenges, we propose a {m}odality-\underline{p}rompted h{e}\underline{t}e{r}ogeneous \underline{g}raph framework \underline{for} \underline{o}mni-modal learning (GTP-4o) that allows unifying representations under various biomedical modalities with potential modality missing.
Specially, we establish a heterogeneous graph embedding~\cite{li2021htd,liu2022towards,li2023novel} to explicitly capture the heterogeneous attributes on both modal features and cross-modal relations.
Then, we design a modality-prompted completion that completes the deficient graph embedding of missing modality through a novel graph prompting module, which generates hallucination nodes to steer the embedding towards the original complete space.
Through the completed graph, we meticulously develop a knowledge-guided hierarchical aggregation that includes a knowledge-derived global meta-path neighbouring to capture the potential heterogeneous neighbors, and a local multi-relation aggregation for the comprehensive interaction of modal information across various heterogeneous relations.
GTP-4o presents the first exploration in learning unified representations from various heterogeneous clinical modalities including genomics, pathological images, cell graphs, and diagnostic texts.
Our contributions can be summarized as follows:

\begin{itemize}
\item{
This paper introduces the new problem of learning unified multimodal representations from various diverse clinical modalities, 
and presents the first effort to embed and relate heterogeneous multimodal features through a graph representation and aggregation.
}
\item{
We propose a modality-prompted completion module to complete the corrupted graph embedding of the missing modality by a graph prompting strategy, which generates hallucination nodes to steer the missing embedding towards the complete representation.
}
\item{
We present a knowledge-guided hierarchical cross-modal aggregation,  employing a global meta-path neighbouring to capture heterogeneous neighbors, and a local multi-relation aggregation module for information interaction across various heterogeneous relations.
}
\item{
Extensive experiments on comprehensive benchmarks of disease diagnosis including 
pathological glioma grading and survival outcome prediction exhibits the efficacy of our method against 
prior state-of-the-arts.
}
\end{itemize}

\section{Related Work} 
\label{sec:related work}
\noindent\textbf{Biomedical Multimodal Learning.}
The utilization of multimodal data has gained significant attention for accurate and comprehensive imaging analysis~\cite{zhou2019review,li2021consistent,liu2022source} and 
diagnosis~\cite{ali2022comprehensive,yang2023mrm,sun2022few}.
For instance, the comprehensive features from pathological images~\cite{zhu2017wsisa,lu2021data}, genomics~\cite{liberzon2015molecular,zeng2021exploration,liu2021tumor}, 
are employed in joint for an accurate cancer-related diagnosis, e.g. glioma grading~\cite{doyle2007automated} and survival analysis~\cite{chen2021multimodalsurvival}.
Meanwhile, with the eGTP-4once of Visual Language Models (VLMs), much efforts have been devoted to enhancing the recognition and analysis ability of vision models by further incorporating the textual information from clinical text reports~\cite{zhang2022contrastivemedtext, zhang2023biomedclip}.
Inspired by these trends, this paper introduces the new problem of learning unified features from various disparate clinical modalities, including genomics~\cite{liberzon2015molecular, chen2021multimodalsurvival}, pathological images~\cite{lu2021data}, cell graphs~\cite{xu2022instancecellgraph0} and text descriptions~\cite{zhang2022contrastivemedtext}.

Handling the modality heterogeneity is critical when integrating multimodal information ~\cite{lipkova2022artificial}.
Early works focus on studying early or late fusion methods~\cite{huang2020fusion,wang2021gpdbn, ramachandram2017deep, joo2021multimodal,richard2022tmi}, which integrates the predictions from individually separated models for the final decision.
However, these methods suffer from either neglecting intra-modality dynamics~\cite{lipkova2022artificial} or failing to fully relating cross-modal information. 
Recent progress in intermediate fusion~\cite{chen2021multimodal,kumar2019co} has shown promise, which  learns uni-modal features and capture cross-modal interactions at the same time, by leveraging the power of cross-modal attention~\cite{chen2021multimodal, zhou2023cross,Xu_2023_ICCV,liu2023efficientvit}.
However, they try to model all potential cross-modal relations with the learned attentions. Different from them, we present to explicitly capture the heterogeneity of modal features and cross-modal relations resorting to a heterogeneous graph space.

\noindent\textbf{Graph Representation in Pathology.}
Graph representation has shown its promise in the field of pathology analysis~\cite{chen2021whole, wang2021online,chan2023histopathology}.
Following previous efforts of Multiple Instance Learning (MIL) that split the high-resolution whole slide pathological images (aka., WSIs) into a bag of instances and pre-define the connective local areas in the Euclidean space, 
recent graph-based methods~\cite{zheng2022graph, chen2021whole, hou2022h,chan2023histopathology} models the interactions among instances flexibility via the graphs topology.
For instance, PatchGCN \cite{chen2021whole} models pathological images with homogeneous graphs, and regress survival data with a graph convolutional neural network (GCN) \cite{gcn}.
GTNMIL \cite{zheng2022graph} is designed as a graph-based MIL using graph transformer networks \cite{yun2019graph}.  
Recent methods \cite{hou2022h, chan2023histopathology} extend the prior practice to handling WSIs with heterogeneous graphs, introducing heterogeneity in each patch by different resolution levels~\cite{hou2022h}, or semantic representations via pretext tasks~\cite{chan2023histopathology}. 
However, these methods only considers heterogeneity in the image modality, while the more challenging multimodal scenario is left to study.
To fill this gap, this paper explores the graph representation in a way more complex setting, i.e., learning from various disparate clinical modalities with significant heterogeneity.

\begin{figure*}[!t]   
\centering	   
\includegraphics[width=\linewidth]{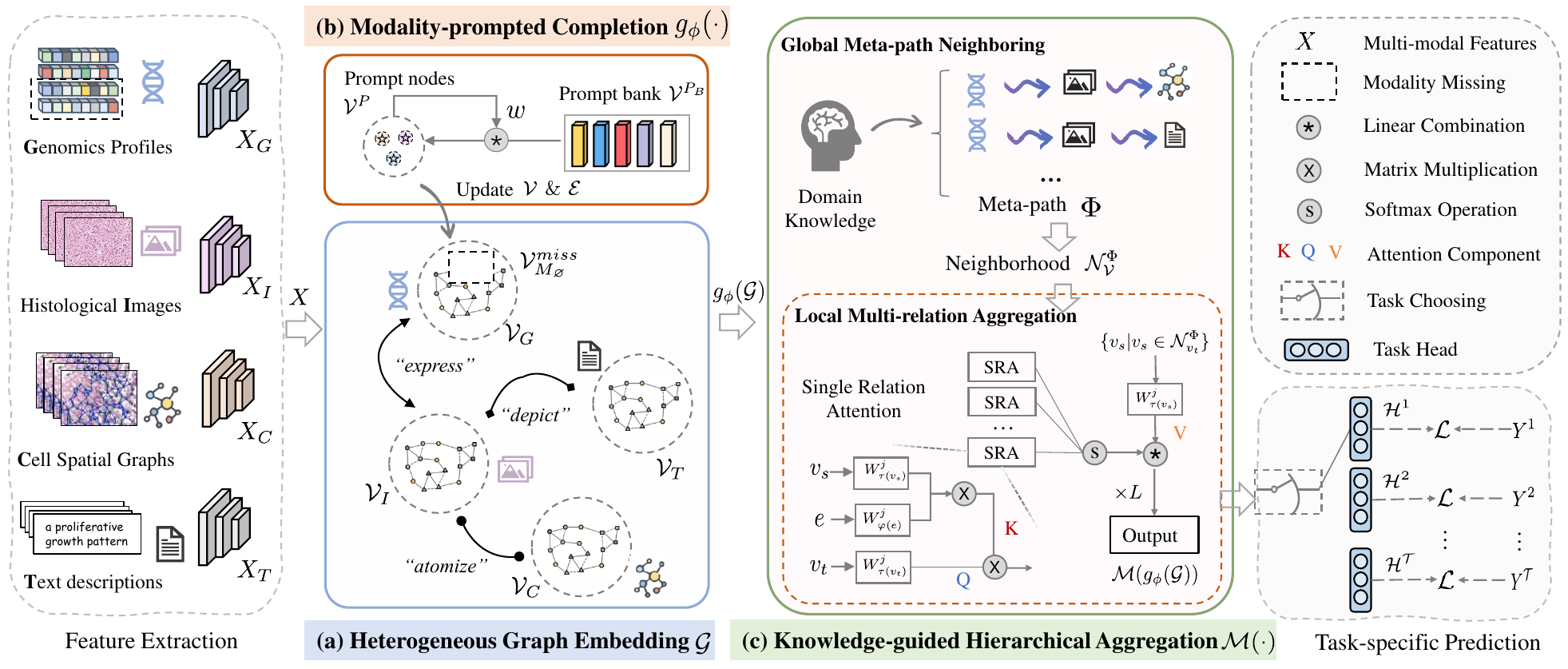}     
\caption{ 
   \textbf{Pipeline Overview of GTP-4o}. 
    We instantiate the omni-modal biomedical features (Sec.~\ref{method1}), and embed them onto \textcolor{Turquoise}{(\textbf{a})} the heterogeneous graph space (Sec.~\ref{method2}).
   Then, we introduce \textcolor{YellowOrange}{(\textbf{b})} the modality-prompted completion via graph prompting to complete the missing embedding (Sec.~\ref{method3}).
   After that, we design \textcolor{ForestGreen}{(\textbf{c})} the knowledge-guided hierarchical aggregation from a global meta-neighbouring to uncover the heterogeneous neighbourhoods and a local multi-relation aggregation to interact features across various heterogeneous relations (Sec.~\ref{method4}).
 }
  \label{fig:framework}    
\end{figure*}

\section{Method}
\label{sec:method}
\noindent\textbf{Overview.}
After performing data processing and feature extraction (Sec.~\ref{method1}), the omni-modal embedding for a patient subject could be represent by a 4-tuples of four modalities, including genomics (G), pathological images (I), cell spatial graphs (C) and diagnostic texts (T), $X=\{X_G, X_I, X_C, X_T\}$, with different number of instances in each modality, while the common dimension $d$.
Then, we establish the heterogeneous graph representation $\mathcal{G}$ by transforming modal features to the graph space (Sec.~\ref{method2}).
After that, the modality-prompted completion is performed, which employs a graph prompting $g_\phi(\cdot)$ to transform the incomplete graphic embedding to a prompted and completed representation $g_\phi(\mathcal{G})$
(Sec.~\ref{method3}).
Afterwards, we conduct a knowledge-guided hierarchical aggregation that is parameterized by $\mathcal{M}$, including a global neighbouring via knowledge-derived meta-paths $\Phi$ , and a local multi-relation aggregation along various heterogeneous relations (Sec.~\ref{method4}).
The final aggregated features $\mathcal{M} \circ g_\phi(\mathcal{G})$ end up with forwarding a task-specific head $\mathcal{H}^\mathcal{T}$ to obtain the specific prediction for task $\mathcal{T}$,
based on which we 
we optimize the network parameters $\mathcal{M}, \mathcal{H}^\mathcal{T}$ and prompt parameters $g_\phi$ \textit{w.r.t.} the task loss $\mathcal{L}$.
(Sec.~\ref{method5}).

\subsection{Data Processing and Feature Extraction}
\label{method1}
\noindent\textbf{Genomic Profiles.}
Following~\cite{richard2022tmi}, the genomic profiles that we use includes Copy Number Variation (CNV), bulk RNA-Seq expressions and mutation status~\cite{richard2022tmi}.
We merge the mutation status and CNV and feed it and RNA-Seq as separate groups of gnomic data into  Self-Normalizing Neural Network (SNN)~\cite{klambauer2017self} 
to get the embedding
$X_G\in \mathbb{R}^{N_G\times d}$, where $N_G$ equals to the number of genomic groups.


\noindent\textbf{Pathological Images.}
Following~\cite{chen2021multimodal,shao2021transmil}, 
we divide WSIs into a series of non-overlapping patches,
employ a ImageNet pretrained ResNet-50 to extract the 
features from each patch, and feed them into a projection layer to obtain $X_I \in\mathbb{R}^{N_I\times d}$, where $N_I$ is the number of patches.

\noindent\textbf{Cell Spatial Graphs.} 
Cell graph representations explicitly capture selected fine-grained features of cells~\cite{marusyk2012intra}.
We utilize the procedure in PathomicFusion~\cite{richard2022tmi} to segment cells for each slide, curate graph topology.
and use a graph convolutional network (GCN)~\cite{gcn} backbone to obtain the aggregated graph embedding, as $X_C \in\mathbb{R}^{N_c\times d}$, where $N_C$ equals to the number of curated cell graphs.

\noindent\textbf{Text Descriptions.}
As no actual medical reports are provided in the used materials, we employ an open-source multimodal Large Language Model (LLM), MiniGPT-4~\cite{zhu2023minigpt} to generate the customized text descriptions for each pathological image\footnote{
We find that another foundation VLM, BLIP-2~\cite{li2023blip} showing effective on curating captions for 3D representation~\cite{xue2023ulip,li2023steganerf,li2024gaussianstego} does not work well for pathological images.}.
The prompt is customized as follows.
\\
\textbf{Prompt:}
\textit{
``This is a pathology slide with glioma cells. Write a caption for this slide based on the following properties: \ding{202} size and shape of cells, \ding{203} color of cells, \ding{204} growth pattern and cellularity of cells,  \ding{205} uclear atypia and pleomorphism of cells, \ding{206} necrosis of cells, \ding{207} microvascular proliferation of cells,  \ding{208} mitotic activity of cells.''
}\\
We obtain several sentences for each slide, like:
\textit{``Some cells show signs of necrosis, with dark spots in the cytoplasm''}.
We take each sentence as a modal instance, and employs a MedBERT~\cite{rasmy2021med} to obtain the embedding
$X_T \in \mathbb{R} ^{N_T\times d}$, where $N_T$ equals to the number of curated sentences.

\subsection{Heterogeneous Graph Embedding}
\label{method2}
With the obtained modal features $X=\{X_G, X_I, X_C, X_T\}$ with the same dimension $d$, 
the heterogeneous graph embedding could be established by a feature to graph transformation.
Formally, a heterogeneous graph space is formulated by  $\mathcal{G}=\{\mathcal{V}, \mathcal{E}, \mathcal{A}, \mathcal{R}\}$, where $\mathcal{V}$ and $\mathcal{E}$ represents the set of entities (i.e., vertices or nodes) and relations (i.e., edges) that has been established in the theory of a classic directed graph.
The further introduced $\mathcal{A}$ and $\mathcal{R}$ represent the attribute set of nodes and edges, respectively, by which we can explicitly define the heterogeneous properties for the features of modal entities and cross-modal relations.
A function $\tau(v)=a\in \mathcal{A}$, is defined to map each node $v$ to an attribute in the set $\mathcal{A}$, according to its modality, 
As a result, the attribute set of nodes can be formulated as $\mathcal{A}=\{G, I, C, T\}$.
Furthermore, the edges $e\in \mathcal{E}$ in the heterogeneous embedding represent the relations from the source nodes $v_s\in{\mathcal{V}_s}$ to the target nodes $v_t\in{\mathcal{V}_t}$\footnote{
For simplicity, index $s$ and $t$ is omitted when cross-modal relations are not involved.}, therefore the attribute of an edge $e:{v_s \to v_t}$ 
is determined is determined by the attribute of the source node $v_s$ and target node $v_t$ as well as their actual semantic relations.
Thus, a function $\varphi(e)=r \in \mathcal{R}$ that maps each edge $e \in \mathcal{E}$ to a specific attribute $r \in  \mathcal{R}$ is introduced.
We formulate this attribute set of relations $\mathcal{R}$ by the prior knowledge of biomedical modalities, 
$\mathcal{R}=\{\textit{``express''}, 
\textit{``depict''}, 
\textit{``atomize''}, 
\textit{``intra-modal''}, 
\}$.  
This set represents the semantic relations of \textit{``express''} between genomics and images, \textit{``depict''} between images and texts, and \textit{``atomize''} between images and cell graphs.
We also model all the relations between 
the \textit{``intra-modal''} instances.
To obtain the initial input graph embedding $\mathcal{V}$, we perform a non-linear projection on the modal features $X$. For each edge, we compute the cosine correlation between the head and tail nodes as its embedding.

\subsection{Modality-prompted Completion}
\label{method3}
The modality-prompted completion aims to adapt the deficient embedding of missing modality by updating its with some prompted entities that could be learnt.
Formally, we introduce a graph prompt operation,  which could be parameterized by $g_\phi$,  transforming a input graph representation $\mathcal{G}$ into $g_\phi(\mathcal{G})$.
Hopefully, it is learnt to transform the missing graph embedding back to its original complete status.

\noindent\textbf{General Prompting.} 
Given the missing modality $M_\varnothing \in \{G, I, C, T\}$, some specific subjects 
have all the instances of that modality missed, such that the representation of them at $M_\varnothing$ is ruined, as $\mathcal{V}_{M_\varnothing}=\{\varnothing\}$.
There are also some patient subjects not affected by the missing, still with the complete data and the representation $\mathcal{V}_{M_\varnothing}$ maintained.
We sample hallucination nodes $v^P\in\mathcal{V}^P$, where $\mathcal{V}^P \in \mathbb{R}^{N_P\times d}$, as a basic prompt scheme for graph completion.
%
We extract the representation prior of the missing modality $M_\varnothing$ by collecting the modality-specific feature from all subjects, except for the subjects with the incomplete data at the missing modality, $\mathcal{V}_{M_\varnothing}$  
The subjects with the incomplete data at the missing modality, i.e., $\mathcal{V}_{M_\varnothing}=\{\varnothing\}$ 
Then we initialize the features of the $N_P$ prompt entities by a Gaussian sampling from the extract modality prior, motivated by the intuition that
the same modality among different subjects share a basically similar distribution.
After initialized, the set of prompt nodes $\mathcal{V}^P$, could be optimized effectively along with the model training.

\noindent\textbf{Entity-dependent Prompting.} 
The introduced prompted entities are agnostic to the context, bringing the risk of yielding sub-optimal results.
To encode the entity-dependent contextual information, we further introduce a prompt bank that contains a set of prompt components, $\mathcal{V}^{P_B}\in \mathbb{R}^{N_B \times {d}}$, where $N_B$ is the number of prompt components.
We take the these components as a series of base prompt, the weights $\boldsymbol{w}$ of which could be obtained in an entity-independent fashion.
That is, we pass each input node $v^P$  through a channel-downscaling linear layer to obtain a compact feature vector, followed by a softmax operation, thus yielding the weights ${w} \in \mathbb{R}^{N_{B}}$,
\begin{equation}
{w} = \texttt{Softmax} \left(W_{d\times N_B}(v^P)\right ),
\end{equation}
where 
$W_{d\times N_P}$ 
denotes the linear projection layer that transforms the feature dimension from $d$ to $N_P$. 
Then we use these weights to modulate the prompt components for each query entity $v^P$, and sum the general prompts and the entity-dependent prompts that perceives the graphic context,
\begin{equation}
    v^P \leftarrow  v^P + \sum_{i=1}^{N_{B}} {w}_i \cdot v^{P_B}_i, 
    \label{eq:promptbank}
\end{equation}
where $v^{P_B}_i$ denotes $i$-th component in the prompt bank. By doing so, the prompted graph embedding could be described with the formulation of graph prompt function $g_\phi(\cdot)$, by which the nodes $\mathcal{V}$ and edges $\mathcal{E}$ of a graph $\mathcal{G}$ to prompt would be transformed as,
\begin{equation}
    \mathcal{V} =  \{\mathcal{V}_{/{M_\varnothing}}, \mathcal{V}^{P} \}, 
\ 
 \mathcal{E} =  \{\mathcal{E}_{/{M_\varnothing}}, 
\underset{\forall\varphi(e)\in\mathcal{R}}{\texttt{EdgeUpdate}}(\mathcal{\mathcal{V}_{/M_\varnothing}},\mathcal{V^P})\}
\end{equation}
where $\mathcal{V}_{/{M_\varnothing}}$ denotes the node embedding at all the modalities except for ${{M_\varnothing}}$ and $\mathcal{E}_{/{M_\varnothing}}$ denotes the edge space when removing all the nodes at the modality ${{M_\varnothing}}$. 
$\underset{\forall\varphi(e)\in\mathcal{R}}{\texttt{EdgeUpdate}}(\cdot, \cdot)$
defines the operation that 
updates the features of edge $e$ between two sets of nodes if the relation can be retrieved in the attribute space of edge $\varphi(e) \in \mathcal{R}$.
Effectively, the graph embedding of the missing modality $M_\varnothing$ are adapted through inserted with the prompt nodes as well as uncovered with some ruined relations. 

\subsection{Knowledge-guided Hierarchical Aggregation} 
\label{method4}
With the completed graph $g_\phi(\mathcal{G})$, 
the knowledge-guided hierarchical aggregation module effectively embeds the knowledge prior into a series of meta-paths, thereby we can search for the global cross-modal heterogeneous neighboring.
With the found neighbouring, the local multi-relation aggregation module is performed across various heterogeneous edges, and the overall hierarchical aggregation~\cite{li2022hierarchical,zhang2021generator} module can be parameterized by a network function $\mathcal{M}(\cdot)$.

\noindent\textbf{Global Meta-path Neighbouring.}
The aggregation of graph information highly depends on the established neighboring rules~\cite{sun2011pathsim,HetGNN2019}, and we design novel meta-paths as global information pathways, allowing for interaction of two heterogeneous entities.
Given the entities $\mathcal{V}$ and meta-paths $\Phi$ in the heterogeneous graph, the neighbors derived from meta-paths for all the entities are uniquely identified~\cite{HetGNN2019}, as $\mathcal{N}^{\Phi}_\mathcal{V}$.
Hence, our insight is to embed the domain knowledge into the formulation of $\Phi$ by considering the semantic relations across the clinical modalities.
Recall that the edge attribute space $\mathcal{R}$ is explicitly defined by the biological relations among modalities, we derive that, with the exception of \textit{``intra-modal''} relations, i.e., the entities at the same modality, all heterogeneous nodes can only interact with the nodes whose attributes are semantically related to themselves in a single-hop propagation~\cite{HetGNN2019,liang2021unsupervised}. 
Following this principle, we adopt a random walking strategy~\cite{codling2008random} to search for the optimal meta-paths from all potential candidates. 
Specially, we randomly start from an entity of one modality, then iterate over all the 
heterogeneous nodes with valid semantic relations in the attribute space, i.e., $\varphi(e_{c}) \in \mathcal{R}$.
Repeating the iterations, we show that an appropriate customization of meta-paths could be 
$\Phi =  \{G \xrightarrow{\textit{``express''}} I \xrightarrow{\textit{``atomize''}} C, \quad
 G\xrightarrow{\textit{``express''}} I \xrightarrow{\textit{``depict''}} T, \quad
C \xrightarrow{\textit{``atomize''}} I \xrightarrow{\textit{``express''}} G, \quad
T \xrightarrow{\textit{``depict''}} I \xrightarrow{\textit{``express''}}G\}$.
Following~\cite{HetGNN2019}, all meta-paths $\Phi$ are formulated with the maximum lengths within two hops.
In practical usage, when querying the neighbourhood $\mathcal{N}^{\Phi}_v$ for an entity $v$, we first project it onto the entity attribute space $\mathcal{A}$ by $\tau(v)$, and then
iterate over all meta-paths in $\Phi$ with a given number of hops $H$.  After that, all reached entities attributes are collected, and the entities of those collected attributes are taken as the neighbours along the meta-paths,
\begin{equation}
    \mathcal{N}^{\Phi}_v = \left\{v' | \tau(v')\in\{ \underset{i\in[1,|\Phi|]}{||}\underset{H}{\texttt{Reach}}({\Phi_i})\} \right\},
\end{equation}
where $\underset{H}{\texttt{Reach}}$ denotes the operation that collects all the reached attributes by walking along a meta-path $\Phi_i$ with $H$ hops. ${||}_{i\in[1,|\Phi|]}$ is the concatenation operator for all the resulted elements.

\noindent\textbf{Local Multi-Relation Aggregation.}
With the derived entity-wise $\mathcal{N}^{\Phi}_v$ neighbours, we perform the information propagation for each target node $v_t\in \mathcal{V}_t$ as a local feature aggregation from all its neighbored source nodes $\mathcal{V}_s$.
To model node-wise interaction~\cite{HGT2020, chan2023histopathology}, we introduce a {M}ulti-{H}ead {A}ttention (MHA) mechanism that models the target node features as \textbf{Q}uery and source node features as \textbf{K}ey and \textbf{V}alue.
We embed the target node $v_t$ and source node $v_s$ by different linear projection layers ${W}_{\tau(v_s)}^j$ and ${W}_{\tau(v_s)}^j$, with each attention head $j$,
\begin{equation}
\begin{aligned}
v^{K,j}_s = & ~{W}_{\tau(v_s)}^j \cdot v_s^{(l-1)}, \quad 
v^{Q,j}_t  ={W}_{\tau(v_t)}^j \cdot v_t^{(l-1)}, \\
& v^{V,j}_s ={W}_{\tau(v_s)}^j \cdot v_s^{(l-1)},
\end{aligned}    
\end{equation}
where $v_{*}^{(l-1)}$ represents the input node feature for node $v \in \mathcal{V}$ from the $(l-1)$-th layer. 
The projection layers are capable of mapping node features from different node attributes to an embedding space that is invariant across node attributes.
%
The features of edges from the $(l-1)$-th layer $e_{v_s\rightarrow v_t}^{(l-1)}$
are also projected by a linear projection layer $W_{\varphi(e)}$, serving as a \textbf{K}ey feature, 
$
e^{K}_{v_s\rightarrow v_t} = W_{\varphi(e)}\cdot e_{v_s\rightarrow v_t}^{(l-1)} 
$. 
Once node embeddings are projected, we calculate the dot-product between the query and key vectors. Besides, we multiply the linearly transformed edge embedding with the similarity score to integrate the edge features into graph $\mathcal{G}$,
\begin{equation}
\begin{gathered}
\texttt{SHA}(e, j)
=\left(
{v}_s^{K,j} \cdot
e^{K}_{v_s\rightarrow v_t} 
\cdot
{v}_t^{Q,j} \right) / \sqrt{d},\\
\end{gathered}    
\end{equation}
where $d$ denotes the dimension of node embeddings, $\texttt{SHA}(e, j)$ represents the attention score of edge $e$ by the Single-Head Attention at $j$-head.
We concatenate the scores obtained from each head and apply a softmax to them,
\begin{equation}
\texttt{SRA}(e)
=\underset{\forall v_s \in 
\mathcal{N}^{\Phi}_{v_t}
}
{
\texttt
{Softmax}}\left( 
\underset{j \in [1,h]}
{||}~\texttt{SHA}
(e, j)
\right), 
\end{equation}
where  $\texttt{SRA}(e)$ represents Single-Relation Attention, providing the final attention score of the edges aggregating all the heads $j \in [1,h]$. $\mathcal{N}^{\Phi}_{v_t}$ is the set of the neighbours to the target node $v_t$.
Then, we perform target-specific aggregation to update the feature of each target node by averaging its neighboring node features. 
For each target node $v_t$, we conduct a softmax operation on all the attention vectors from its neighboring nodes and then aggregate the information of all neighboring source nodes of $v_t$ together. The updated node features $v_t^{(l)}$ for $\mathcal{G}^{(l)}$ can be represented as,
\begin{equation}
v^{(l)}_t = \underset{\forall v_s \in \mathcal{N}_{v_t}^\Phi}{\bigoplus} \left (
\underset{j \in [1,h]}
{||}~
\left(v_s^{V,j}
\cdot \texttt{SRA}(e)\right)
 \right),    
\end{equation}
where $\oplus$ is an aggregation operator, e.g., mean aggregation. The updated graph $\mathcal{G}^{(l)}$ is returned as the output of the $l$-th layer. Such operation is scalable by using $L$ layers of aggregation.
We further introduce modality-specific pooling for all nodes within the modality to obtain the prototype features for all modalities.
Then the graph level feature can be further determined by a mean readout layer~\cite{HetGNN2019,chan2023histopathology}.

\subsection{Overall Optimization}
\label{method5}

The aggregated multimodal representation could be obtained  by $\mathcal{M}\circ g_\phi (\mathcal{G}) $ from the heterogeneous graph $\mathcal{G}(\cdot)$, modality-wise graph prompting $g_\phi(\cdot)$ and knowledge-guided hierarchical aggregation $\mathcal{M}(\cdot)$.
While the diverse diagnostic tasks including the glioma grading (a classification task) and the survival prediction (a integration prediction task) that may differ in the formulation, it is shown that they could be transformed to a uniform supervised learning fashion after some manipulations of task head~\cite{richard2022tmi}.
Formally, the task-specific task head $\mathcal{H}^\mathcal{T}$ for the task $\mathcal{T}$ is introduced, with the task label denoted by $y^\mathcal{T}$,
\begin{equation}
\min_{\mathcal{M}, \mathcal{H}, \phi} \mathbb{E}_{\mathcal{G}}~\mathcal{L}\left(\mathcal{H}^\mathcal{T}\circ\mathcal{M}\circ\phi (\mathcal{G}), y^\mathcal{T}\right).
\end{equation}
 where $\mathcal{L}$ denotes the loss function, which could be implemented by a NLL (negative log-likelihood) loss.

\section{Experiments}
\subsection{Datasets and Settings}

\noindent\textbf{Datasets.} 
We evaluate our method using data from The Cancer Genome Atlas (TCGA)~\cite{TCGA}
, a public database that includes genomic and clinical data from thousands of cancer patients.
We select the datasets of Glioblastoma \& Lower Grade Glioma (GBMLGG) and Kidney Renal Clear Cell Carcinoma (KIRC).
For TCGA-GBMLGG, following~\cite{richard2022tmi}, 
we use ROIs from diagnostic slides and apply sparse stain normalization~\cite{richard2022tmi} to match all images to a standard H\&E histology image, creating a total of 1505 images for 769 patients, with WHO grading labels from G2 to G4.
We curate 80 CNA and 240 RNA-Seq genomic features for each patient.
Note that there are 40\% of the patients with inherently \underline{actual missing} RNA-Seq data.
For the KIRC dataset, we use manually extracted 512 × 512 ROIs from diagnostic whole slide images for 417 patients in CCRCC, yielding 1251 images total that are similarly normalized with stain normalization. We pair these images with 117 CNV and 240 RNA-Seq genomic features. There are grading labels by Fuhrman Grading from G1 to G4.

\noindent\textbf{Evaluation.} 
For each cancer dataset, we perform 5-fold cross-validation and report the average test performance.
Different metrics are leveraged for specific evaluation tasks, pathological glioma grading with Area Under the Curve (AUC) and  Accuracy (ACC), and survival outcome prediction with concordance index (C-Index).
Due to inherent missing of partly genetic modality in GBMLGG, we deploy the framework directly without modifying the data.While for KIRC with the complete modality data, we simulate the same situations in GBMLGG, randomly dropping RNA-Seq data with 40\% subjects over the whole dataset.
To ensure the consistent missing cases at training and test, we constrain the proportion of incomplete subjects in the training and test splits to be equal when producing the five-fold validation.
In order to explore the missing issues under more modalities and missing ratios, we also perform experiments under simulated missing settings in Sec.~\ref{Further}.

\noindent\textbf{Implementation.}
The framework is optimized by the Adam optimizer, with a learning rate of $1\times10^{-3}$ and a weight decay of $1\times10^{-5}$ for the graph aggregation $\mathcal{M}$ and task head $\mathcal{\mathcal{H}}$, over 150 epochs with early stopping. 
We adopt a smaller learning rate of $2\times10^{-4}$ specially for optimizing prompt nodes $\mathcal{V}_P$ and the prompt bank components $\mathcal{V}^{P_B}$ in graph prompt function $g_\phi(\cdot)$.
All the multimodal instances of a patient subject are jointly fed to the networks to get the final
Data augmentations are performed on the training graphs, which involve randomly dropping edges and nodes, and adding Gaussian noise to the node and edge features~\cite{li2022knowledge,xu2022afsc}. The dropout ratio of each dropout layer is selected as 0.2. 
Regarding the hyper-parameters, we have the number of prompt nodes and prompt bank components in  Eq.~\ref{eq:promptbank} with $N_P=5$ and $N_B=5$, and the dimension of input graph representation as $d=512$. 

\begin{table}[!t]
  \centering
  \caption{
  \textbf{Performance Comparison.}
   We report results on four TCGA benchmarks, using various modality combinations of \underline{G}enomics, \underline{I}mages, \underline{C}ell graphs and \underline{T}exts.
  }
\vspace{-0.5em}
  \resizebox{0.9\columnwidth}{!}{
    \begin{tabular}{lcccccccccc}
    \toprule
    \multicolumn{1}{l}{\multirow{2}[4]{*}{Methods}} & \multicolumn{4}{c}{Modality}  & \multicolumn{4}{c}{Glioma Grading (AUC/ACC)} & \multicolumn{2}{c}{Survival\,Pred.\,(C-Idx)} \\
    \cmidrule(lr){2-5} \cmidrule(lr){6-9}    \cmidrule(lr){10-11} 
    
& G     & I     & C     & T     & \multicolumn{2}{c}{GBMLGG} & \multicolumn{2}{c}{KIRC} & GBMLGG & KIRC \\
    \midrule
    SNN~\cite{klambauer2017self}   & \cmark  & \xmark & \xmark & \xmark & 0.8527  & 0.6583  & 0.8100  & 0.7790  & 0.7974  & 0.6639  \\
    SNNTrans~\cite{shao2021transmil} &  \cmark    & \xmark & \xmark & \xmark &  0.8678  & 0.6725  & 0.8084  & 0.7755  & 0.7970  & 0.6671  \\
    \midrule
    AttMIL~\cite{ilse2018attention} &    \xmark   &   \cmark     &   \xmark    &   \xmark    & 0.9063  & 0.7533  & 0.8252  & 0.7803  & 0.7908  & 0.6850  \\
    TransMIL~\cite{shao2021transmil} &    \xmark   &  \cmark      &    \xmark   &   \xmark    & 0.9149  & 0.7683  & 0.8295  & 0.7899  & 0.8017  & 0.6876  \\
    PatchGCN~\cite{chen2021whole} &   \xmark    &  \cmark     &  \xmark     & \xmark      & 0.8802  & 0.7429  & 0.8288  & 0.7896  & 0.7806  & 0.6795  \\
    GTNMIL~\cite{zheng2022graph} &     \xmark  &  \cmark    &  \xmark     &  \xmark     & 0.9225  & 0.7966  & 0.8323  & 0.7980  & 0.8162  & 0.6953  \\
    HEAT~\cite{chan2023histopathology}   &  \xmark     &   \cmark    &    \xmark   &  \xmark     & 0.9289  & 0.8057  & 0.8300  & 0.7961  & 0.8223  & 0.7059  \\
    \midrule
    Pathomic~\cite{richard2022tmi} &   \cmark      &  \cmark      &   \xmark    &    \xmark   & 0.9172  & 0.7618  & 0.8295  & 0.7899  & 0.8101  & 0.7152  \\
    Porpoise~\cite{chen2022pan}  &   \cmark      &  \cmark      &   \xmark    &    \xmark   &0.9199  & 0.7789  & 0.8278  & 0.7800  & 0.8179 & 0.7179  \\
    MCAT~\cite{chen2021multimodal}  &   \cmark      &  \cmark      &   \xmark    &    \xmark   & 0.9288  & 0.7929  & 0.8352  & 0.7957  & 0.8274 & 0.7235  \\
    TransFusion~\cite{zhou2023cross} &   \cmark      &  \cmark      &   \xmark    &    \xmark   & 0.9209  & 0.7815  & 0.8299  & 0.7910  & 0.8251 & 0.7230  \\
    GTP-4o (Ours) &   \cmark      &  \cmark      &   \xmark    &    \xmark   & \cellcolor{top3}0.9256  & \cellcolor{top3}0.8036  & 0.8349  & 0.7985  & \cellcolor{top3}0.8296  & \cellcolor{top3}0.7273  \\
    \midrule
    Pathomic~\cite{richard2022tmi} & \cmark & \cmark  & \cmark &  \xmark & 0.9195  & 0.7674  & 0.8280  & 0.7889  & 0.8199  & 0.7211  \\
    TransFusion~\cite{zhou2023cross} & \cmark & \cmark  & \cmark &  \xmark &      0.9225  & 0.7952  & 0.8318  & 0.7973  & 0.8283  & 0.7260  \\
    GTP-4o (Ours)  & \cmark & \cmark  & \cmark &  \xmark & \cellcolor{top2}0.9336  & \cellcolor{top2}0.8068  & \cellcolor{top2}0.8331  & \cellcolor{top2}0.8021  & \cellcolor{top2}0.8329  & \cellcolor{top2}0.7315  \\
    \midrule
    TransFusion~\cite{zhou2023cross}   & \cmark & \cmark  & \cmark &  \cmark &   0.9245  & 0.7986  & 0.8325  & \cellcolor{top3}0.7990  & 0.8296  & 0.7289  \\
    GTP-4o (Ours) 
     & \cmark & \cmark  & \cmark &  \cmark &   
      \cellcolor{top1}0.9389 & \cellcolor{top1}0.8126  & \cellcolor{top1}0.8416  &\cellcolor{top1}0.8068  & \cellcolor{top1}0.8351  & \cellcolor{top1}0.7336  \\
    \bottomrule
    \end{tabular}%
    }
        \vspace{-1em}
  \label{tab1}%
\end{table}%

\subsection{Comparison with State-of-the-arts}
We compare the proposed method against several SOTA methods in Tab.~\ref{tab1}. For fair comparison, we apply identical settings for all experiments, and use the official code of compared works to deploy on our tasks when necessary.

\noindent\textbf{Unimodal Models.}
Existing methods to analyze genomic data and pathological images are introduced. 
For genomic data, we employ {SNN}~\cite{klambauer2017self} for survival outcome prediction in the TCGA~\cite{richard2022tmi,chen2021multimodal}, and {SNNTrans}~\cite{klambauer2017self,shao2021transmil} that incorporates SNN as the feature extractor and TransMIL~\cite{shao2021transmil} for a global aggregation.
For pathological images, we report the results of the SOTA MIL methods including the transformer-based models: {AttnMIL}~\cite{ilse2018attention}, 
{TransMIL}~\cite{shao2021transmil},
and the graph-based models {PatchGCN}~\cite{chen2021whole}, {GTNMIL}~\cite{zheng2022graph}, {HEAT}~\cite{chan2023histopathology}.
It appears that using multimodal data consistently improves the performance under various metrics.

\noindent\textbf{Multimodal Models.}
We compare the SOTA multimodal methods including {Pathomic}~\cite{richard2022tmi}, {Porpoise}~\cite{chen2022pan} and {MCAT}~\cite{chen2021multimodal}, which only focus on extracting complementary multimodal information from the genomics and pathological images.
It appears that there is a gain in using multimodal complementary information for various diagnostic tasks.
Furthermore, as the study of extending to more modalities is still unexplored, we compare our GTP-4o with other baselines by extend existing work {Pathomic}~\cite{richard2022tmi} with the cell graph modality, and also compare with a simple baseline {TransFusion}~\cite{zhou2023cross} which concentrates the intra-modal representations learned by uni-modal models~\cite{shao2021transmil}. From the table, the proposed GTP-4o exhibits the obvious performance improvement under most of the usage of different biomedical modalities.

\begin{table}[t]
  \centering
  \caption{
  \textbf{Ablation of GTP-4o Variants.}
  Results on TCGA-GBMLGG benchmarks over tasks of Glioma Grading (GG.) and Survival Prediction (SP.) are reported.}
\vspace{-6pt}
\resizebox{0.9\columnwidth}{!}{
    \begin{tabular}{llccc}
    \toprule
 Components & {Variants} & GG.\,(AUC)  & GG.\,(AUC)  & SP.\,(C-Idx) \\
    \midrule
   \multirow{2}[2]{*}{Graph Representation} &  No \textit{Heterogeneous Embedding}  & 0.9232  & 0.8030  & 0.8168
 \\
   & No \textit{Heterogeneous Relation} & 0.9259  & 0.8048  & 0.8201
 \\
    \midrule
   \multirow{3}[2]{*}{Modality Completion} &   No \textit{Completion (Zero-init Missing)}  & 0.9087  & 0.7875  & 0.7946
 \\
 & No \textit{Completion (Drop Missing)} & {}0.9288  & {}0.8061  & {}0.8233
 \\
 & No \textit{Prompt Bank} &  \cellcolor{top3}0.9275  & \cellcolor{top2}0.8081  & 0.8280
 \\
 \midrule
 \multirow{2}[2]{*}{Hierarchical Aggregation} & No \textit{Aggregation (Plain Mean)} & 

 {}0.9329  & {}0.8067  & \cellcolor{top3}0.8311
 \\
  & No \textit{Knowledge Guidance} & \cellcolor{top2}0.9350  & \cellcolor{top3}0.8071  & \cellcolor{top2}0.8342
 \\
    \midrule
 Full Model &  The Proposed GTP-4o   & \cellcolor{top1}0.9389  & \cellcolor{top1}0.8126  & \cellcolor{top1}0.8416
 \\
    \bottomrule
    \end{tabular}%
    }
    \vspace{-10pt}
  \label{tab2}%
\end{table}%

\subsection{Further Results}
\label{Further}
\noindent\textbf{Ablation Studies.}
The effect of removing each component of GTP-4o is presented in Tab.~\ref{tab2}.
\textit{No Heterogeneous Embedding} removes all the heterogeneous properties in the embedding such that it degrades to a simple homogeneous graph.
\textit{No Heterogeneous Relation} removes the heterogeneous properties of edges while maintaining the diverse attributes among the node features.
\textit{No Completion (Zero-init Missing)} handles the missing modality without using the proposed graph prompt completion while sets the features of the missing modality to zero values.
\textit{No Completion (Drop Missing)} directly drops the modality data in all patients if it occurs missing for some patients.
\textit{No Aggregation (Plain Mean)} removes the knowledge-guided aggregation while performs the plain mean aggregation among the $k$-NN heterogeneous neighbours ($k=15$).
\textit{No Knowledge Guidance} removes the knowledge guidance for aggregation while uses the random meta-paths.
Our ablation study results confirm the pivotal roles of our designs in the overall performance and effectiveness of the model.

\begin{figure}[t]   
	\centering	   
\includegraphics[width=\linewidth]{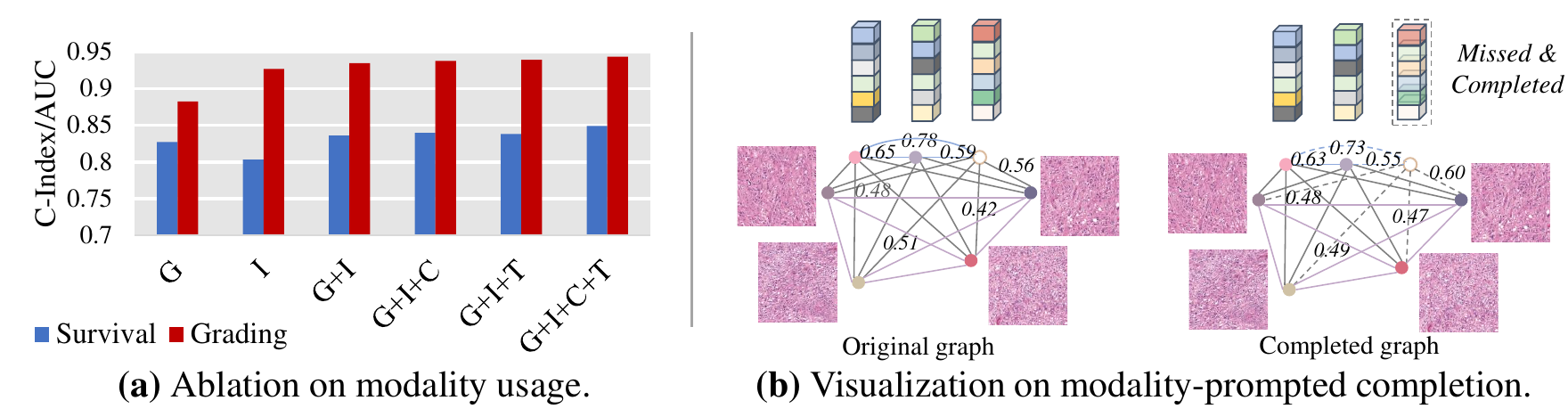}  
\vspace{-12pt}
\caption{
\textbf{(a)} \textbf{Analysis of Modality Usage.} We provide the results of GTP-4o by using either \underline{G}enes, \underline{I}mages, \underline{C}ell graphs, \underline{T}exts, or their combinations, on benchmarks of survival prediction (C-Index) and glioma grading (AUC).
\textbf{(b)}  \textbf{Analysis of Modality-prompted Completion}. We compare the relation pattern (similarity) in the original graph and the graph that is first removed specific instances then completed. 
}
  \label{fig:exp1}   
\vspace{-5pt}
\end{figure}

\noindent\textbf{Impact of Modality Usage.}
Fig.~\ref{fig:exp1}(\textbf{a}) shows the impact of using various combinations of modalities by GTP-4o. 
For the case of single modality, it is observed that each modality has its advantage for a specific task, as the relative performance of using only genes and only images is opposite for the tasks of glioma grading and survival analysis.
We can also see that when more modalities like cell graphs and text descriptions are introduced, the performance of the model is improved on both two tasks.
This suggests that GTP-4o is
not only capable of generalizing to the various combinations of medical modality usage, but also able to
deliver superior performance in terms of AUC (for glioma grading) and C-Index (for survival prediction).  

\begin{figure}[t]   
\centering	   
\includegraphics[width=\linewidth]{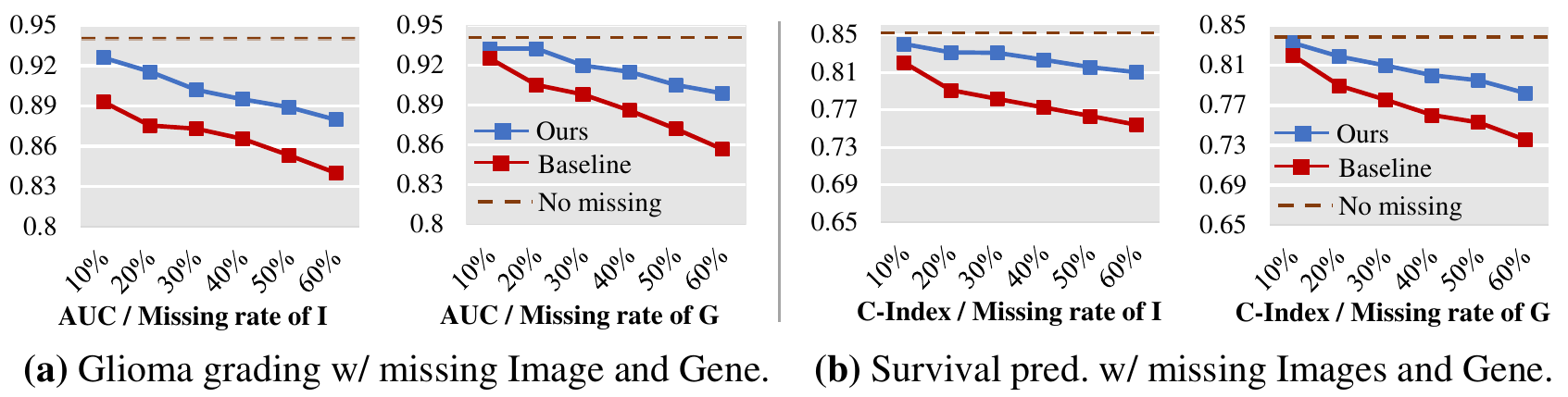}     
\vspace{-1.5em}
\caption{ 
\textbf{Analysis of Modality Missing.} We study the results of \textbf{(a)} glioma grading and \textbf{(b)} survival prediction with the various missing ratios of \underline{I}mages and \underline{G}enes.
We compare the full framework of Ours and the version without our completion (baseline). 
 }
 \vspace{-1em}
  \label{fig:exp2}    
\end{figure}

\noindent\textbf{Impact of Modality Missing and Completion.}
To validate the effect of graph prompting, as shown in Fig.~\ref{fig:exp1}(\textbf{b}), we pick a non-missing case (\textit{TCGA-02-0006}), and
compare two versions, including the graph built on original full instances and the completed graph with arbitrary missing instances, with the edges labelled with cosine similarity.
We can observe the
similar relation patterns of completed graph with the real
one, implying the biological validity of the proposed completion.
Besides, we explore more missing settings by applying simulated missing on the image and genomics modalities on TCGA-GBMLGG benchmarks.
Note that since there is actual missing RNA genomics in GBMLGG dataset, the simulated missing of gene modality is applied only on Non-RNA profiles.
Fig.~\ref{fig:exp2} depicts the performance of GTP-4o over the baseline without graph-prompted completion under various missing ratios, confirming that the effectiveness of the proposed completion under various modality missing setting.

\noindent\textbf{Limitation and Future Works.}
The current deployment is limited by the fact that no real-world clinical text reports are available for the datasets, thus we have to generate synthetic text descriptions by LLMs, probably bringing some data noise.
Another limitation is that some additional modalities such as tabular data, are not considered in this paper, which could serve as future works.

\vspace{-3pt}
\section{Conclusion}
Increasing biomedical multimodal data provides not only opportunities for accurate and comprehensive diagnosis but also challenges for learning against the modality heterogeneity as well as the missingness issues.
This study presents GTP-4o, which signifies a pioneering exploration into learning unified representations from various clinical modalities via the graph theory, exhibiting the robustness to heterogeneous modalities.
Unlike prior methods, GTP-4o explores capturing explicit relations via a heterogeneous graph embedding.
A novel graph prompting is proposed to complete deficient graph representations of missing modalities, and a hierarchical multimodal aggregation employs a global meta-path prior to guide the local aggregation across various heterogeneous relations. 
Extensive experiments demonstrate the efficacy of GTP-4o on disease diagnosis.

\bibliographystyle{splncs04}
\bibliography{main}
\end{document}